\documentclass[letterpaper]{article} 
\usepackage{aaai25}  
\usepackage{times}  
\usepackage{helvet}  
\usepackage{courier}  
\usepackage[hyphens]{url}  
\usepackage{graphicx} 
\usepackage{natbib}  
\usepackage{caption} 
\usepackage{algorithm}
\usepackage{algorithmic}

\usepackage{algorithm}
\usepackage{algorithmic}
\usepackage{amsmath}

\usepackage{amssymb}
\usepackage{bm}
\usepackage{natbib}

\usepackage{amsthm}

\usepackage{stackengine}
\usepackage{subfigure}
\usepackage{enumitem}
\usepackage{caption}
\usepackage{graphicx}
\usepackage{multirow}
\usepackage{float} 
\usepackage{booktabs}

\newtheorem{lemma}{Lemma}
\newtheorem{definition}{Definition}

\newtheorem{thm}{Theorem}

\newcommand{\bx}{\mathbf{x}}

\newcommand{\Xcal}{\mathcal{X}}

\newcommand{\Hcal}{\mathcal{H}}
\newcommand{\Rcal}{\mathcal{R}}

\newcommand{\norm}[1]{\|#1\|}

\renewcommand{\eqref}[1]{Eq.~(\ref{#1})}

\renewenvironment{proof}[1][Proof: ]{\noindent \textit{#1}}{\qed\medskip}

\newenvironment{nospaceflalign*}
 {\setlength{\abovedisplayskip}{10pt}\setlength{\belowdisplayskip}{10pt}%
  \csname flalign*\endcsname}
 {\csname endflalign*\endcsname\ignorespacesafterend}

\usepackage{newfloat}
\usepackage{listings}
\DeclareCaptionStyle{ruled}{labelfont=normalfont,labelsep=colon,strut=off} 
\floatstyle{ruled}
\newfloat{listing}{tb}{lst}{}
\floatname{listing}{Listing}
\title{Pareto Set Learning for Multi-Objective Reinforcement Learning}
\author{
    Erlong Liu,
    Yu-Chang Wu,
    Xiaobin Huang,
    Chengrui Gao,\\
    Ren-Jian Wang,
    Ke Xue,
    Chao Qian
}
\affiliations{
    National Key Laboratory for Novel Software Technology, Nanjing University, Nanjing 210023, China\\
    School of Artificial Intelligence, Nanjing University, Nanjing 210023, China\\
    \{liuel, wuyc, huangxb, gaocr, wangrj, xuek, qianc\}@lamda.nju.edu.cn
}

\begin{document}

\maketitle

\begin{abstract}
Multi-objective decision-making problems have emerged in numerous real-world scenarios, such as video games, navigation and robotics. Considering the clear advantages of Reinforcement Learning (RL) in optimizing decision-making processes, researchers have delved into the development of Multi-Objective RL (MORL) methods for solving multi-objective decision problems. However, previous methods either cannot obtain the entire Pareto front, or employ only a single policy network for all the preferences over multiple objectives, which may not produce personalized solutions for each preference. To address these limitations, we propose a novel decomposition-based framework for MORL, Pareto Set Learning for MORL (PSL-MORL), that harnesses the generation capability of hypernetwork to produce the parameters of the policy network for each decomposition weight, generating relatively distinct policies for various scalarized subproblems with high efficiency. PSL-MORL is a general framework, which is compatible for any RL algorithm. The theoretical result guarantees the superiority of the model capacity of PSL-MORL and the optimality of the obtained policy network. Through extensive experiments on diverse benchmarks, we demonstrate the effectiveness of PSL-MORL in achieving dense coverage of the Pareto front, significantly outperforming state-of-the-art MORL methods in the hypervolume and sparsity indicators.
\end{abstract}

\section{Introduction}

Reinforcement Learning~(RL)~\citep{sutton2018reinforcement} has become a prevalent technique for addressing decision-making and control problems in a wide range of scenarios, including video games~\citep{mnih2015human}, navigation~\citep{yang2019generalized}, robotics~\citep{schulman2017proximal}, dynamic algorithm configuration~\citep{xue2022multi}, and financial cloud services~\citep{yang2024reducing}. In standard RL, the main objective is obtaining an optimal policy that achieves the best expected returns. However, in many real-world scenarios, spanning from robotics, where the balance between speed and energy efficiency is paramount~\citep{todorov2012mujoco}, to smart systems design, where the trade-offs between competing metrics must be dynamically managed~\citep{azzouz2018handling}, the problems may involve multiple conflicting objectives. Contrary to standard RL, Multi-Objective RL (MORL) aims to generate an optimal policy tailored to each preference over multiple objectives, constructing a diverse set of optimal policies that collectively form a Pareto set.

Traditional MORL methods transform the multi-dimensional objective space into a scalar representation~\citep{mannor2001steering} through the static assignment of weights to each objective, achieving a trade-off between multiple objectives. Such approaches, however, struggle with objectives of varying magnitudes and require extensive domain knowledge to set the appropriate weights/preferences~\citep{amodei2016concrete}. Moreover, they inherently limit the solution to a singular policy for a predetermined weight, disregarding the demand for multiple trade-off solutions in real-world scenarios. When extending such approaches to find the dense Pareto set, the inefficiency of repetitively retraining the policy is unacceptable~\citep{parisi2014policy}. Recent studies have proposed several multi-policy methods, which simultaneously learn a set of policies over multiple preferences using a single network~\citep{parisi2017manifold,chen2019meta,xu2020prediction,basaklar2023pdmorl}. However, these methods either cannot construct a continuous Pareto front, or will suffer from the curse of dimensionality and poor performance when confronting with the conflicting objectives due to the reason that they only obtain a single network for all the preferences.

To address these limitations, in this work, we integrate Pareto Set Learning~(PSL)~\citep{lin2022mobo, lin2022nmoco} within the context of MORL, utilizing the specialty of PSL in covering the whole preference space. 
Considering that the solutions to MORL problems are generated by the policy networks, we propose to employ hypernetwork~\citep{ha2017hypernetworks} as the PSL model to generate a continuum of Pareto-optimal policy parameters using a decomposition-based method, and thus generate the relatively personalized policy network for each preference.
Since the trained hypernetwork can generate policy parameters for every possible weight without any additional cost (e.g., evaluation and fine-tuning), our proposed framework can dynamically align with user-specified preferences~\citep{abels2019dynamic}. It is worth noting that our method is a general MORL framework, which can be equipped with any standard RL algorithm. We also perform theoretical analysis, proving the superiority of the model capacity of PSL-MORL by using Rademacher complexity~\cite{golowich18a}, and the optimality of the obtained policy network by using Banach Fixed Point Theorem~\citep{banach18stoc}.

We conduct experiments to demonstrate the effectiveness of PSL-MORL across several benchmarks, containing the Fruit Tree Navigation (FTN) task with discrete state-action spaces~\citep{yang2019generalized} and multiple continuous state-action control tasks MO-MuJoCo~\citep{todorov2012mujoco,xu2020prediction}. Since PSL-MORL is a general framework, we employ Double Deep Q-Network (DDQN)~\citep{van2016deep} for discrete environments and Twin Delayed Deep Deterministic Policy Gradient (TD3)~\citep{fujimoto2018addressing} for continuous environments as the lower-level RL algorithms. 
The experimental results demonstrate that our method is superior to state-of-the-art MORL methods, and the ablation study also verifies the effectiveness of the parameter fusion technique employed by PSL-MORL. 

Our contributions are three-fold:
\begin{enumerate}
    \item We propose a novel MORL method called PSL-MORL, which, to the best of our knowledge, is the first MORL method that covers all preferences over multiple objectives and outputs a personalized policy network for each preference. PSL-MORL is a general framework that can be integrated with any single-objective RL algorithm.
    \item The theoretical results guarantee the superiority of the model capacity of PSL-MORL over previous methods, as well as the optimality of the generated policy network.
    \item The experiment results show that PSL-MORL outperforms all the other MORL baselines both in the hypervolume and sparsity metrics.
\end{enumerate}

\section{Background}
\subsection{MORL Problem}
A multi-objective sequential decision problem can be characterized as a Multi-Objective Markov Decision Process (MOMDP)~\citep{hayes2022review}, represented by a tuple $\langle S, A, T, \bm \gamma, \mu, \mathbf{R} \rangle$, where $S$ is the state space, $A$ is the action space, $T \colon S \times A \times S \to \left[ 0, 1 \right]$ is a probabilistic transition function, $\bm \gamma=\left[\gamma_1, \gamma_2, ..., \gamma_m \right] \in [0, 1]^m$ is a discount factor vector, $\mu \colon S \to [0,1]$ is a probability distribution over initial states, and $\mathbf{R}=\left[r_1, r_2, ..., r_m \right]^{\mathrm{T}} \colon S \times A \times S \to \mathbb{R}^m$ is a vector-valued reward function, specifying the immediate reward for each of the considered $m$ ($m\geq 2$) objectives. When $m=1$, it is specialized to a single-objective MDP. 

In MOMDPs, an agent behaves according to a policy $\pi_{\theta}\in\Pi$, where $\Pi$ is the set of all possible policies. 
A policy is a mapping ${\pi_{\theta}}: S \to A$, which selects an action according to a certain probability distribution for any given state. For brevity, we use $\pi$ to denote $\pi_{\theta}$. The performance of the policy is measured by the associated vector of expected returns, i.e., $\mathbb{F}(\pi)=J^{\pi}=\left[J_1^\pi, J_2^\pi, ..., J_m^\pi \right]^{\mathrm{T}}$ with
\begin{equation}
    J_i^\pi = \mathbb{E} \left[ \sum\limits^\infty_{k=0} \gamma_i^k r_i(s_k,a_k,s_{k+1}) \:|\: \pi, \mu \right].
\end{equation}
The MORL problem is then formulated as 
\begin{equation}
    \max_{\pi} \mathbb{F}(\pi)=\max_{\pi}\left[ J_1^\pi,J_2^\pi,...,J_m^\pi \right]^{\mathrm{T}}.
\end{equation}

In single-objective settings, the expected return offers a complete ordering over the policy space, i.e., for any two policies $\pi$ and $\pi^\prime$, $J^\pi$ will either be greater than, equal to, or lower than $J^{\pi^\prime}$. Thus, it is sufficient to find an optimal policy $\pi^*$ that maximizes the expected cumulative discounted reward. However, in MORL, there exists no single optimal policy that can maximize all the objectives simultaneously, and we need the following Pareto concepts~\citep{qian2013analysis,qian2019maximizing,zhou2019evolutionary}.

\begin{definition}[Pareto Dominance]\label{def-pareto-dominance} For $\pi,\pi^\prime \in \Pi$, $\pi$ is said to weakly dominate $\pi^\prime$ ($\pi \succeq \pi^\prime$) if and only if $\forall i \in \{1,...,m\}, J_i^\pi \geq J_i^{\pi^\prime}$; $\pi$ is said to dominate $\pi^\prime$ ($\pi \succ \pi^\prime$) if and only if $\pi \succeq \pi^\prime$ and $\exists j \in \{1,...,m\}, J_j^\pi > J_j^{\pi^\prime}$.
\end{definition}

\begin{definition}[Pareto Optimality] A solution $\pi^{\ast} \in \Pi$ is Pareto optimal if $\nexists \,\hat \pi \in \Pi$ such that $\hat \pi \succ \pi^{\ast}$.
\end{definition}

\begin{definition}[Pareto Set/Front] The set of all Pareto optimal solutions is called the Pareto set, and the image of the Pareto set in the objective space is called the Pareto front.
\end{definition}

Given a fixed \emph{utility function} $u$ mapping the multi-objective expected returns of a policy to a scalar value, i.e., $u({J^\pi}, \bm{\omega})= \bm{\omega}^{\mathrm{T}} {J^\pi}$, where $\bm \omega=[\omega_1,\dots,\omega_m]^{\mathrm{T}},\omega_i\geq 0$ and $\sum_{i}\omega_i=1$, the problem will be reduced to a single-objective decision-making problem, which can be solved using standard RL methods to generate a Pareto optimal solution corresponding to the preference $\bm{\omega}$. However, in many complex scenarios, the size of the Pareto front is extremely large. Thus, the goal of MORL is often to efficiently obtain a good approximation of the Pareto front.

\subsection{MORL Algorithms}
Previous MORL approaches can be mainly divided into single-policy, multi-policy, and meta-policy approaches. Single-policy approaches transform a multi-objective problem into a single-objective problem by combining rewards into a single scalar reward using a utility function, and then use standard RL methods to maximize the scalarized return~\citep{roijers2013survey}. However, these approaches require domain-specific knowledge and predefined preferences~\citep{van2013scalarized,abdolmaleki2020distributional}. Multi-policy approaches aim to obtain a good approximated Pareto set, and the most widely used approach is running a single-policy algorithm repeatedly over various preferences~\citep{roijers2014linear, mossalam2016multi, zuluaga2016varepsilon}. A more recent study proposes an efficient evolutionary learning algorithm to update a population of policies simultaneously in each run to improve the approximation to the Pareto front~\citep{xu2020prediction}. However, it suffers from the low efficiency of acquiring dense Pareto optimal solutions and handling a large number of objectives~\citep{pirotta2015multi,parisi2017manifold,chen2019meta}.

Recent studies have proposed multi-objective Q-learning approaches that simultaneously learn a set of policies over multiple preferences, using a single network that takes preferences as inputs and uses vectorized value function updates~\citep{abels2019dynamic,yang2019generalized}. Another orthogonal approach frames MORL as a meta-learning problem using a task distribution given by a distribution over the preference space~\citep{chen2019meta}. It first trains a meta-policy to approximate the Pareto front implicitly, and then obtains the Pareto optimal solution of a given preference by fine-tuning the meta-policy with a few gradient updates. But it often fails because the meta-policy may not be optimal, leading to the ultimate policy being suboptimal. PD-MORL~\citep{basaklar2023pdmorl} is a recently proposed method that utilizes a universal neural network to obtain the entire Pareto front. However, these approaches obtain a single policy network for all the trade-off weights, making it challenging to simultaneously address all scalarized subproblems. Compared to previous methods, our proposed PSL-MORL method generates a policy network for each preference, thereby alleviating conflicts and enhancing the overall quality of the solution set.

Multi-Objective Evolutionary Algorithms (MOEAs), inspired by natural evolution, are another sort of algorithms that can solve MORL problems. They are competitive approaches in solving complicated multi-objective optimization problems~\citep{qian2015constrained,qian2015pareto,qian2015subset,qian2017subset,hong2018scalable} and can be applied to some continuous control problems~\citep{fcs-QianY21}. They obtain a high quality solution set by iteratively using crossover, mutation, and selection operators. For instance, MOEA/D (MOEA based on Decomposition)~\citep{moead} is one of the most common MOEA algorithms, which partitions a multi-objective optimization problem into a series of single-objective optimization or less complex multi-objective optimization components. It then employs a heuristic search approach to concurrently and collaboratively refine these easier optimization sub-tasks. However, MOEAs face the challenges in scaling up with the dimensionality of parameters and are not ideally suited for the training of large-scale neural networks in RL.

\subsection{Pareto Set Learning and Hypernetworks} 
Previous multi-objective optimization algorithms can only attain a finite Pareto set/front, while PSL~\citep{lin2020controllable,navon2021learning} encompasses the utilization of a neural network that is tailored to user preferences, enabling the acquisition of the complete Pareto set/front. A standard PSL problem is formulated as $\min_{\theta\in\Theta}g_{WS}(x=h_\theta(\lambda)\mid \lambda), \forall \lambda \in \Lambda$, where the function $g_{WS}$ is a weighted sum of the objective functions, i.e., $g_{WS}=\sum_{i=1}^m \lambda_i f_i(x)$, $\Lambda$ is the preference distribution, and $h_{\theta}$ is the Pareto set model and is often implemented as a hypernetwork. If we take the expectation over all preferences, it turns out to be $$\min_{\theta\in\Theta}\mathbb{E}_{\lambda\sim\Lambda}g_{WS}(x=h_\theta(\lambda)\mid \lambda),$$ which can be computed using stochastic optimization, i.e., $\min_{\theta\in\Theta} \sum_i g_{WS}(x=h_\theta(\lambda_i)\mid \lambda_i ), \{ \lambda_i \} \sim \Lambda$. PSL has been successfully applied to multi-objective Bayesian optimization~\citep{lin2022mobo,lu2024you}, multi-objective neural combinatorial optimization~\citep{lin2022nmoco}, and drug design~\citep{jain2023multiobjective}. Recently,~\citep{zhang2024hypervolume} provided a novel geometric perspective for PSL and recognized its equivalence to hypervolume maximization.

The concept of hypernetwork was first introduced in~\citep{ha2017hypernetworks}. Instead of directly training a primary network to derive the optimized parameters, the hypernetwork takes a different approach: The weights within the primary network remain static, while the adaptable parameters of the hypernetwork are refined through backpropagation and a preferred update mechanism. A key advantage of hypernetworks lies in their dynamic model generation capabilities, which allow for creating customized models for a given input. This is achieved through a single learnable network that provides a versatile and efficient approach to generate a diverse array of models tailored to specific needs.

Considering that PSL is a strong tool in covering the whole preference space and meanwhile hypernetwork has a strong ability in enhancing the model representation ability, we adopt PSL for MORL and implement it as a hypernetwork to generate the parameters for the main policy network.

\section{Pareto Set Learning for MORL}

\begin{figure*}[t!]
  \centering
  \includegraphics[scale=0.30]{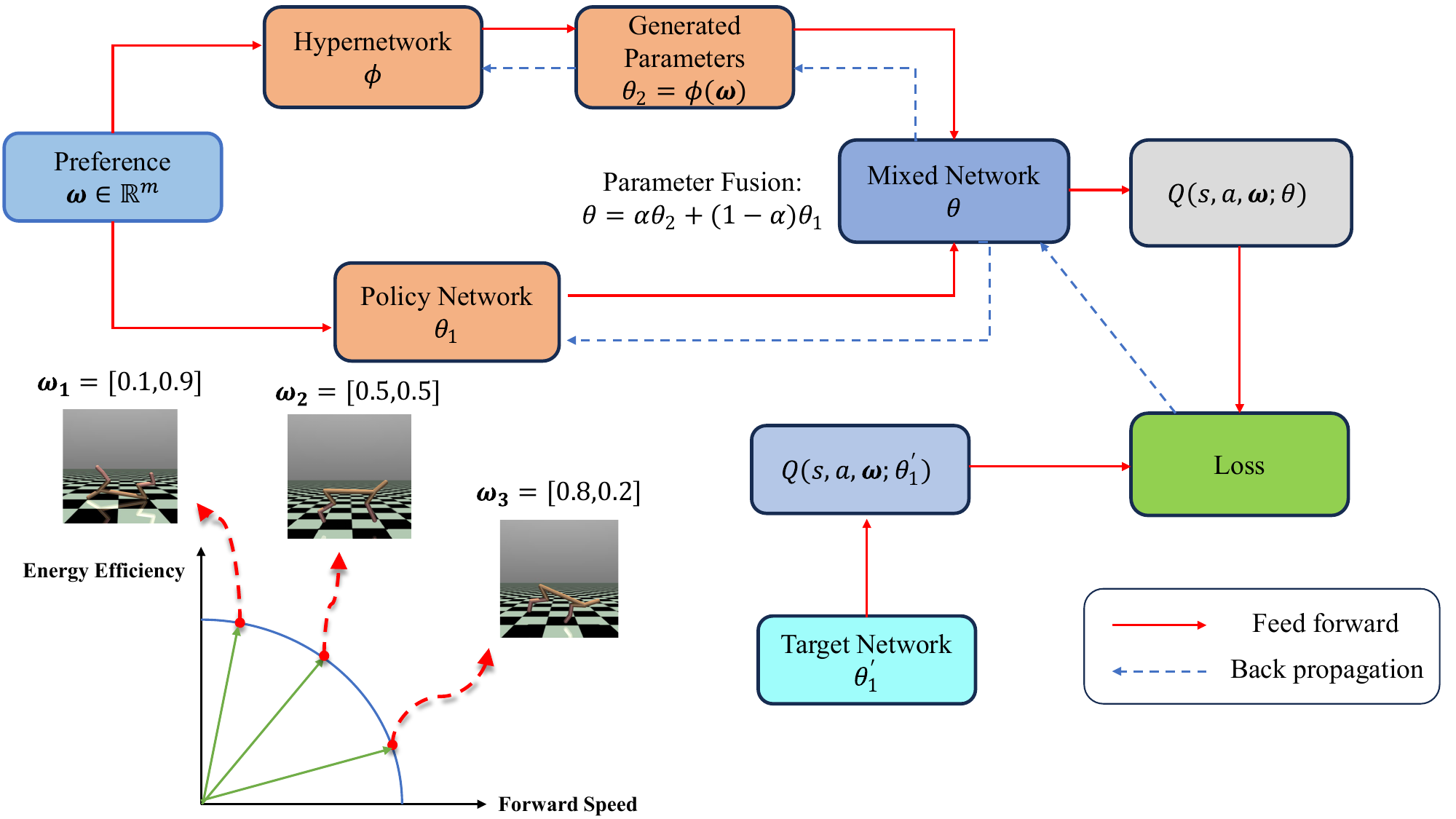}
  \caption{Illustration of the proposed PSL-MORL method. The ultimate parameters of the policy network is composed of two portions, one is the parameters $\theta_2=\phi(\bm{\omega})$ generated by the hypernetwork, and the other is the initial parameters $\theta_1$ of the policy network. The left part is the input of the whole PSL-MORL, i.e., the preference randomly sampled from the uniform distribution. The middle part is the two portions, and the right part is the parameter fusion. Through mixing the parameters, we can get the final policy network and derive the loss to update our hypernetwork and the policy network. The output is the optimal parameters for our hypernetwork and the policy network parameters $\theta_1$.} 
  \label{fig:framework}
\end{figure*}

In this section, we introduce the Pareto Set Learning method for MORL~(PSL-MORL), an efficient framework for approximating the whole Pareto front of MORL. The overall workflow of PSL-MORL is intuitively illustrated in Figure~\ref{fig:framework}. PSL-MORL consists of two primary modules: a hypernetwork $\phi$ and a main policy network. The hypernetwork, when given a preference vector $\bm{\omega}$ as input, generates the parameters for the main policy network. This policy network is then used to control an agent in response to various states. During each training episode, a random preference vector $\bm{\omega}$ is drawn from a uniform distribution. After inputting the preference into the hypernetwork, the agent is run in the environment using the generated policy network parameters $\phi(\omega)$. The hypernetwork parameters, along with the target network, are subsequently updated through gradient backpropagation of the RL loss. In practice, to enhance training stability, we co-train another policy network with the hypernetwork using the parameter fusion technique~\citep{ortiz2023magnitude}.

In the following, we will first introduce the details of the general PSL-MORL framework. Then, we give the instantiations of PSL-MORL with DDQN, and prove the optimality of the generated policy network. Finally, we theoretically analyze the model capacity of PSL-MORL, and prove its superiority over the state-of-the-art method PD-MORL~\cite{basaklar2023pdmorl}, which thus provides theoretical justification for its greater ability of generating more personalized policies for various preference vectors.

\begin{algorithm}[t!]
  \caption{PSL-MORL}
    \label{alg_psl_morl}
    \textbf{Input:}
        Preference distribution $\Lambda$, environment $\mathcal{E}$, number $E$ of episodes, number $K$ of weights per episode, replay buffer $\mathcal{D}$, batch size $N$\\
    \textbf{Output:}
        Hypernetwork parameters $\phi$ and primary policy network parameters $\theta_1$ \\\vspace{-1em}
    \begin{algorithmic}[1]
    \STATE Initialize the hypernetwork parameters $\phi$ and policy network parameters $\theta_1$; 
    \STATE Initialize replay buffer $\mathcal{D}$;
        \FOR{episode $e = 1$ to $E$}
          \FOR{each step in the episode}
            \STATE // Parallel child processes for $k \in \{1,2,...,K\}$
            \STATE $\bm{\omega}_k \sim \textbf{SamplePreference}(\Lambda)$;
          \STATE Generate parameters of the policy network by hypernetwork, i.e., $\theta_2 =\phi(\bm{\omega}_k)$;
          \STATE Obtain the mixed $\theta$ using the \textbf{Parameter Fusion} method on $\theta_1$ and $\theta_2$;
            \STATE Observe state $s$; select action $a$ by the policy~$\pi_{\theta}$;
            \STATE Execute action $a$ in environment $\mathcal{E}$, and observe reward $\bm{r}$ and next state $s'$;
            \STATE Store transition $(s, a, \bm{r}, s', d, \bm{\omega}_k)$ in $\mathcal{D}$;
            \STATE // Main process
            \STATE Sample $N$ transitions from $\mathcal{D}$;
            \STATE Update $\phi$ and $\theta_1$ by conducting a single-objective RL algorithm to maximize the scalarized return by weight $\bm{\omega}_i, i\in\{1,2,...,N\}$
          \ENDFOR
        \ENDFOR	
    \end{algorithmic}
\end{algorithm}

\subsection{Details of PSL-MORL}\label{subsec:details}

As shown in Algorithm~\ref{alg_psl_morl}, PSL-MORL takes the preference distribution $\Lambda$, fundamental elements of RL (i.e., environment $\mathcal{E}$, replay buffer $\mathcal{D}$, number $E$ of episodes, and batch size $N$), and number $K$ of weights per episode as input, and outputs the optimized hypernetwork parameters $\phi$ and the primary policy network parameters $\theta_1$. In this work, we use a simple MLP,  $\phi(\bm{\omega})=\text{MLP}(\bm{\omega}\mid \phi)$, as our hypernetwork to generate the policy network parameters conditioned on the preference. The detailed structure of the MLP model can be found in Appendix~A in the full version. 

The objective of training hypernetwork for Pareto set approximation is to maximize the expected scalarized return of MORL, that is, $\max_{\phi} \mathbb{E}_{\bm{\omega}\sim \Lambda}[u(J^{\pi_{\phi(\bm{\omega})}}, \bm{\omega})]$,
where $\pi_{\phi(\bm{\omega})}$ is a policy parameterized by $\phi(\bm{\omega})$, and $u$ is the linearized utility function. The optimization can be based on a single-objective RL algorithm. We divide the optimization process into two phases: the parallel child process and the main process. First, we sample multiple weights from the distribution $\Lambda$ and use the policies generated by the hypernetwork $\phi$ to collect trajectories. These trajectories, along with the corresponding weights, are then stored in the buffer. In the second phase, a single-objective RL algorithm is applied to optimize the utility function $u(J^{\pi_{\phi(\bm{\omega})}}, \bm{\omega})$ for each sampled weight. Note that PSL-MORL is compatible with any single-objective RL algorithm.

To be more specific, PSL-MORL first initializes the hypernetwork parameters $\phi$, policy network parameters $\theta_1$, and experience replay buffer $\mathcal{D}$ in lines~1--2 of Algorithm~\ref{alg_psl_morl}. For each step in each episode, we randomly sample $K$ preferences $\{\bm{\omega}_1,...,\bm{\omega}_K \} \sim \Lambda$ in line~6 for the parallel child process. In line~7, we utilize the hypernetwork to obtain the generated policy parameters $\theta_2$, and get the mixed policy parameters $\theta = (1-\alpha)\theta_1 + \alpha\theta_2$ using the parameter fusion technique (which will be presented in detail in the next paragraph) in line~8.
After that, we use the policy $\pi_{\theta}$ to update the environment and store the transitions $(s, a, \bm{r}, s', d, \bm{\omega}_k)$ to $\mathcal{D}$ in lines~9--11. In the main process (lines~12--14), we randomly sample $N$ transitions from the buffer $\mathcal{D}$, and use an RL algorithm (e.g., DDQN and TD3) to update the hypernetwork $\phi$ and the original policy network $\theta_1$.

\paragraph{Parameter Fusion for Stable Training} As observed in~\citep{ortiz2023magnitude}, the performance of hypernetwork prediction has large fluctuations and is highly sensitive to the norm of the input vector, thus leading to unstable training. To address this issue and stabilize the hypernetwork training, we design a parameter fusion technique to blend the parameters predicted by the hypernetwork and the main policy network parameters, i.e., $$\theta=(1-\alpha)\cdot\theta_1 + \alpha\cdot\phi(\bm{\omega}),$$ where $\theta_1$ denotes the main policy network parameters that are simultaneously optimized with the hypernetwork $\phi$, and $\alpha$ controls the proportionality of parameter magnitudes. In this work, we set $\alpha$ to a proper value for different benchmarks, considering the mixing of the generated parameters by the hypernetwork and the parameters from the primary policy network. Different settings of $\alpha$ will be empirically compared in Appendix~A.6.

\subsection{Instantiation of PSL-MORL with DDQN}\label{subsec:instantiation}
In this subsection, we introduce an instantiation of PSL-MORL by taking DDQN~\citep{van2016deep} as the single-objective RL algorithm. The pseudo code is shown in Appendix~A.1 due to space limitation. The chosen replay buffer is the Hindsight Experience Replay (HER)~\citep{andrychowicz2017hindsight}, which is used to overcome the bias towards over-represented preferences. Inspired by~\citep{basaklar2023pdmorl}, we adopt the idea of dividing the preference space into several subspaces to improve the sample efficiency of the algorithm and using the multi-dimensional preference interpolator $I(\bm{\omega})$ to align the generated solution with the preference vector.

PSL-MORL with DDQN first initializes the hypernetwork and HER. The parallel child processes are the same as that in Algorithm~\ref{alg_psl_morl}. After that, it operates on a batch of preferences. For each preference, the multi-dimensional interpolator $I(\bm{\omega})$ is used to obtain the interpolated preference $\bm w_p$, and a modified Q-learning update scheme (with the cosine similarity term $S_c$ proposed in~\citep{basaklar2023pdmorl}) is used to calculate the target value. The empirical expected loss is then obtained by calculating the DDQN loss for each preference. Finally, an optimizer (e.g., SGD) is employed to update the parameters of both the hypernetwork and target network.

The following theorem~\citep{yang2019generalized, basaklar2023pdmorl} shows that, the $\mathbf{Q}$-network space of the policy is a complete metric space, and the existence of a contraction mapping $\mathcal{C}$ ensures that $\mathbf{Q}^*$ under the optimal policy of any preference is the unique fixed point of this contraction mapping. In other words, 
for PSL-MORL using DDQN as the basis RL algorithm, the generated policy network  can converge to the optimal policy under any preference $\bm \omega$.

\begin{thm}
~\citep{yang2019generalized, basaklar2023pdmorl}
\label{theorem:main}
$(\mathcal{Q},d)$ is a complete metric space, where 
\begin{align*}
&\forall \mathbf{Q}, \mathbf{Q}^\prime \in \mathcal{Q}, d(\mathbf{Q},\mathbf{Q}^\prime) := \\&\qquad \sup_{\substack{s \in \mathcal{S} , a \in \mathcal{A} , \bm{\omega} \in \Lambda}}
|\bm{\omega}^{\mathrm{T}}(\mathbf{Q}(s,a,\bm{\omega})-\mathbf{Q}^\prime(s,a,\bm{\omega}))|.
\end{align*}
Let $\mathbf{Q}^* \in \mathcal{Q}$ be the optimal multi-objective value function, such that it takes multi-objective Q-value corresponding to the supremum of expected discounted rewards under a policy $\pi$. If $\mathcal{C} : \mathcal{Q} \rightarrow \mathcal{Q}$ is a contraction on $\mathcal{Q}$ with modulus $\gamma$, then $\mathbf{Q^*}$ is a unique fixed point such that $\mathbf{Q^*} =\mathcal{C}(\mathbf{Q^*})$. 
\end{thm}

In the experiments, we also adopt PSL-MORL with TD3~\citep{dankwa2019twin}, whose details are shown in Appendix~A.2. Here, we use the hypernetwork to generate only the parameters of actor network and also conduct the parameter fusion technique. Similar to~\citep{basaklar2023pdmorl}, we include the directional angle term to both the actor and critic loss function, which depicts the alignment relationship between the preference vectors and the Q-values. 
 
\subsection{Theoretical Analysis for PSL-MORL}\label{subsec:theoreticAnalysis}

In this subsection, we theoretically analyze the model representation ability of PSL-MORL, which is characterized by the Rademacher complexity in Definition~\ref{definition:Radamacher}. A model with higher Rademacher complexity will have a larger model capacity~\citep{Vapnik00SLT, Zhu09HumanRademacher,Shai14UnderstandingML,Neyshabur19}.

\begin{definition}(Rademacher Complexity~\citep{golowich18a})
\label{definition:Radamacher}
Given a real-valued function class $\Hcal$ and a set of data 
points $\bx_1,\ldots,\bx_n\in \Xcal$, the (empirical) Rademacher complexity $\hat{\Rcal}_n(\Hcal)$ is defined as 
\begin{equation}\label{eq:raddef}
\hat{\Rcal}_n(\Hcal) = 
\mathbb{E}_{\boldsymbol{\epsilon}}\left[\sup_{h\in\Hcal}\frac{1}{n}\sum_{i=1}^{n}
\varepsilon_i h(\bx_i)\right],
\end{equation}
where $\boldsymbol{\varepsilon}=[\varepsilon_1,\ldots,\varepsilon_n]$ is a 
vector 
uniformly distributed in $\{-1,+1\}^n$.
\end{definition}

We prove in Theorem~\ref{thm:rademacher} that PSL-MORL has a higher Rademacher complexity (thus a better model capacity) than PD-MORL~\citep{basaklar2023pdmorl}, the most competitive baseline among all the MORL algorithms. This implies a greater ability of PSL-MORL to provide more personalized policies for various preference vectors, as will also be shown in the experiments. The theoretical analysis relies on the common assumption that the activation function $\sigma$ of a neural network satisfies the 1-Lipschitz property and is positive homogeneous.

\begin{thm}
\label{thm:rademacher}
PSL-MORL has a better model capacity compared to PD-MORL, i.e., $\hat{\mathcal{R}}_2 > \hat{\mathcal{R}}_1$, where $\hat{\mathcal{R}}_2$ and $\hat{\mathcal{R}}_1$ denote the Rademacher complexity of PSL-MORL and PD-MORL, respectively.
\end{thm}

The proof details of Theorem~\ref{thm:rademacher} are provided in Appendix~B.3, and we only provide a proof sketch here. By inequality scaling, we first establish a lower bound $R_2$ for the Rademacher complexity $\hat{\mathcal{R}}_2$ of the hypernetwork MLP used in PSL-MORL, and an upper bound $R_1$ for the Rademacher complexity $\hat{\mathcal{R}}_1$ of the MLP used in PD-MORL. Subsequently, we prove that the ratio $R_2/R_1$ has a lower bound greater than $1$. Then we have $\hat{\mathcal{R}}_2/\hat{\mathcal{R}}_1\geq R_2/R_1>1$, implying $\hat{\mathcal{R}}_2>\hat{\mathcal{R}}_1$.

\section{Experiments}

In this section, we will give the experimental settings and results. The experiments aim to answer the following two research questions:
(1) How is the approximation performance of PSL-MORL compared to other state-of-the-art MORL algorithms? 
(2) How does the parameter fusion technique help PSL-MORL find better policies?
Due to space limitation, some details are shown in Appendix~A.

\subsection{Experimental Settings}
To examine the performance of PSL-MORL, we conduct experiments on two popular MORL benchmarks: a continuous benchmark {MO-MuJoCo}~\citep{xu2020prediction} and a discrete benchmark {Fruit Tree Navigation (FTN)}~\citep{yang2019generalized}. MO-MuJoCo is a popular MORL benchmark based on the MoJocCo physics simulation environment, consisting of several continuous control tasks. We conduct experiments under five different environments, including MO-HalfCheetah-v2, MO-Hopper-v2, MO-Ant-v2, MO-Swimmer-v2, and MO-Walker-v2. The number of objectives is two for all environments in MO-MuJoCo. FTN is a discrete MORL benchmark with six objectives, whose goal is to navigate the tree to harvest fruit to optimize six nutritional values on specific preferences. Full details of the benchmarks are provided in Appendix~A.4. 

To evaluate the performance of PSL-MORL, we compare the following seven state-of-the-art methods: {PG-MORL}~\citep{xu2020prediction}, {PD-MORL}~\citep{basaklar2023pdmorl}, {Radial Algorithm (RA)}~\citep{parisi2014policy}, {Pareto Front Adaption (PFA)}~\citep{parisi2014policy}, {MOEA/D}~\citep{moead}, {RANDOM}~\citep{xu2020prediction}, and {META}~\citep{chen2019meta}. Please refer to Appendix~A.5 for details.

An MORL algorithm aims to find a set of policies to approximate the underlying optimal Pareto set and Pareto front. Following previous works~\citep{chen2019meta,xu2020prediction, liu21pgmeta, basaklar2023pdmorl}, we use two metrics: (i) Hypervolume and (ii) Sparsity~\citep{hayes2022review}, to compare the quality of the set of policies obtained by different methods.

\noindent \textbf{Hypervolume (HV)}
Let $Q$ be an approximated Pareto front in an $m$-dimensional objective space and contain $M$ solutions, more precisely, the objective vectors of $M$ solutions. Let $\mathbf{r}_1 \in \mathbb{R}^m$ be a reference point. The hypervolume indicator is defined as:
\begin{equation}\label{eq:hv}
  \text{HV}(Q) := \alpha(H(Q,\mathbf{r}_1)),
  \nonumber
\end{equation}
\noindent where $H(Q,\mathbf{r}_1)=\{\bm{w} \in\mathbb{R}^m |~\exists~1\leq j \leq M,  \mathbf{r}_1 \preceq \bm{w} \preceq Q_j\}$ with $Q_j$ being the $j$-th solution in $Q$, and $\preceq$ is the weak Pareto dominance operator as defined in Definition~\ref{def-pareto-dominance}. Note that $\alpha(\cdot)$ is the Lebesgue measure with $\alpha(H(Q,\mathbf{r}_1)) = \int_{\mathbb{R}^m} 1_{H(Q,\mathbf{r}_1)}(\bm{w})d\bm{w}$, where 
$1_{H(Q,\mathbf{r}_1)}$ is the indicator function of ${H(Q,\mathbf{r}_1)}$.

\noindent \textbf{Sparsity (SP)} 
For an approximated Pareto front $Q$, sparsity is defined as:
\begin{equation}\label{eq:sp}
    \text{Sparsity}(Q) := \frac{1}{M-1} \sum_{k=1}^{m} \sum_{j=1}^{M-1} (Q^{k}_{j} - Q^{k}_{{j+1}})^2,
    \nonumber
\end{equation}
where for each objective, the solutions are sorted in descending order according their values on this objective, and $Q^k_j$ denotes the $j$-th largest value on the $k$-th objective. Note that if there is only one solution in $Q$, we set the sparsity metric to be N/A.

The hypervolume metric can measure the convergence performance of the approximated Pareto front. However, it cannot fully describe the density of the approximated Pareto front. 
A dense policy set is always favored for better Pareto
approximation~\citep{chen2019meta, xu2020prediction, liu21pgmeta, basaklar2023pdmorl}. Therefore, it is necessary to consider the sparsity metric, which measures the density of the approximated Pareto front in a multi-objective control problem. As a larger hypervolume implies a better approximated Pareto set and a lower sparsity indicates a denser approximated Pareto set, the goal is to obtain an approximated Pareto set with a high hypervolume metric and meanwhile a low sparsity metric. Note that hypervolume and sparsity are the most frequently used metrics to measure the performance of MORL algorithms~\citep{chen2019meta, xu2020prediction, liu21pgmeta, basaklar2023pdmorl}.

\subsection{Results on MORL Benchmarks with Continuous State-Action Spaces}\label{subsec-continuous benchmark}
We first evaluate the methods on the popular multi-objective continuous control benchmark MO-MuJoCo. We run each method with six different random seeds, consistent with previous works~\citep{xu2020prediction, basaklar2023pdmorl}. We do not compare with Envelope Q-Learning~\cite{yang2019generalized} on MO-MuJoCo since it cannot be applied to continuous action spaces. For a fair comparison, we use the same seeds and hyperparameters for other baseline algorithms. As shown in Table~\ref{tab:cont_benchmark}, PSL-MORL achieves the best average rank for both metrics, demonstrating the high quality of the obtained Pareto set. PD-MORL is the runner-up, performing worse than PSL-MORL which may be due to the limited capacity and diversity of its single policy model. PG-MORL involves a large optimization space for training multiple policies simultaneously, achieving the third average rank. RANDOM performs worse than PG-MORL, because it does not consider the influence of the choice of weights. Due to the fixed weights and the low efficiency of training separately, RA doesn't have a good performance on the hypervolume metric.
PFA also performs poorly since it trains the policy separately and the transfer may not be optimal. The low sample efficiency of the gradient-free optimization in MOEA/D leads to its poor performance. META has the worst hypervolume, because it can only obtain a sub-optimal policy.

\begin{table*}[h!]
\centering
\caption{Performance comparison between PSL-MORL and other state-of-the-art algorithms on the popular multi-objective continuous control benchmark MO-MuJoCo in terms of hypervolume and sparsity. We run each algorithm with six random seeds and report the mean value of the two metrics. Bold numbers are the best in each row. The reference points for hypervolume calculation are set to $(0,0)$.
}
\label{tab:cont_benchmark}
\resizebox{0.85\linewidth}{!}{
\begin{tabular}{c|c|cccccccc}
\toprule
 &
  Metrics &
  RA &
  PFA &
  MOEA/D &
  RANDOM &
  PG-MORL &
  META &
  PD-MORL &
  \textbf{PSL-MORL} \\ 
  \midrule
\multirow{2}{*}{MO-Walker2d-v2} &
  HV $(\times10^6)$ &
  $4.82$ &
  $4.16$ &
  $4.44$ &
  $4.11$ &
  $4.82$ &
  $2.10$ &
  $\textbf{5.41}$ &
  $5.36$ \\
 &
  SP $(\times10^4)$ &
  $0.04$ &
  $0.37$ &
  $1.28$ &
  $0.07$ &
  $0.04$ &
  $2.10$ &
  $0.03$ &
  $\textbf{0.01}$ \\
  \midrule
\multirow{2}{*}{MO-HalfCheetah-v2} &
  HV $(\times10^6)$ &
  $5.66$ &
  $5.75$ &
  $5.61$ &
  $5.69$ &
  $5.77$ &
  $5.18$ &
  $5.89$ &
  $\textbf{5.92}$ \\ 
 &
  SP $(\times10^3)$ &
  $15.87$ &
  $3.81$ &
  $16.96$ &
  $1.09$ &
  $0.44$ &
  $2.13$ &
  $0.49$ &
  $\textbf{0.16}$ \\
  \midrule
\multirow{2}{*}{MO-Ant-v2} &
  HV $(\times10^6)$ &
  $5.98$ &
  $6.23$ &
  $6.28$ &
  $5.54$ &
  $6.35$ &
  $2.40$ &
  $7.48$ &
  $\textbf{8.63}$ \\ 
 &
  SP $(\times10^4)$ &
  $5.50$ &
  $1.56$ &
  $1.97$ &
  $1.13$ &
  $\textbf{0.37}$ &
  $1.56$ &
  $0.78$ &
  $0.68$ \\
  \midrule
\multirow{2}{*}{MO-Swimmer-v2} &
  HV $(\times10^4)$ &
  $2.33$ &
  $2.35$ &
  $2.42$ &
  $2.38$ &
  $2.57$ &
  $1.23$ &
  $3.21$ &
  $\textbf{3.22}$ \\ 
 &
  SP $(\times10^1)$ &
  $4.43$ &
  $2.49$ &
  $5.64$ &
  $1.94$ &
  $0.99$ &
  $2.44$ &
  $0.57$ &
  $\textbf{0.42}$ \\
  \midrule
\multirow{2}{*}{MO-Hopper-v2} &
  HV $(\times10^7)$ &
  $1.96$ &
  $1.90$ &
  $\textbf{2.03}$ &
  $1.88$ &
  $2.02$ &
  $1.25$ &
  $1.88$ &
  $1.95$ \\ 
 &
  SP $(\times10^4)$ &
  $5.99$ &
  $3.96$ &
  $2.73$ &
  $1.20$ &
  $0.50$ &
  $4.84$ &
  $\textbf{0.30}$ &
  $0.68$ \\
  \midrule
\multirow{2}{*}{Average Rank} &
  HV &
  $5.1$ &
  $5.2$ &
  $4.2$ &
  $6.1$ &
  $2.9$ &
  $8.0$ &
  $2.7$ &
  $\textbf{1.8}$ \\ 
 &
  SP &
  $6.7$ &
  $5.9$ &
  $7.0$ &
  $4.2$ &
  $2.3$ &
  $6.1$ &
  $2.2$ &
  $\textbf{1.6}$ \\
  \bottomrule

\end{tabular}
}
\end{table*}

\begin{table*}[h!]
\centering
\caption{Ablation study on the parameter fusion technique on the multi-objective continuous control benchmarks. We run all algorithms on each
problem with 6 different random seeds and report the mean value of the hypervolume and sparsity metrics. The bold number is the best in each column. The reference points for hypervolume calculation are set to $(0,0)$.}
\label{tab:ablation}
\resizebox{0.90\linewidth}{!}{
\begin{tabular}{@{}lcccccccccc@{}}
\toprule
 & \multicolumn{2}{c}{MO-Walker2d-v2} & \multicolumn{2}{c}{MO-HalfCheetah-v2} & \multicolumn{2}{c}{MO-Ant-v2} & \multicolumn{2}{c}{MO-Swimmer-v2} & \multicolumn{2}{c}{MO-Hopper-v2} \\ \midrule
 & \multicolumn{1}{c}{HV $(\times10^6)$} & \multicolumn{1}{c}{SP $(\times10^4)$} & \multicolumn{1}{c}{HV $(\times10^6)$} & \multicolumn{1}{c}{SP $(\times10^3)$} & \multicolumn{1}{c}{HV $(\times10^6)$} & \multicolumn{1}{c}{SP $(\times10^4)$} & \multicolumn{1}{c}{HV $(\times10^4)$} & \multicolumn{1}{c}{SP $(\times10)$} & \multicolumn{1}{c}{HV $(\times10^7)$} & \multicolumn{1}{c}{SP $(\times10^4)$}  \\\midrule

 \textbf{PSL-MORL(Ours)} & $\mathbf{5.36}$ & $\mathbf{0.01}$ & $\mathbf{5.92}$ & $\mathbf{0.16}$ & $\mathbf{8.63}$ & $\mathbf{0.68}$ & $\mathbf{3.22}$ & $\mathbf{0.42}$ & $\mathbf{1.95}$ & $\mathbf{0.68}$  \\
\textbf{PSL-MORL-gen}
& $1.47$   & $0.55$& $4.65$  &  $0.75$ & $1.39$  &  $1.95$    &  $2.11$   &   $3.64$  &  $0.83$ & $6.88$    \\
\textbf{PSL-MORL-add}
& $2.63$ & $1.26$ & $5.62$ & $1.56$ & $4.27$ & $2.13$ & $2.67$ & $3.68$ & $1.28$ & $8.19$ \\

\bottomrule
\end{tabular}
}
\end{table*}

\begin{table}[h!]
\caption{Comparison on the multi-objective discrete benchmark FTN in terms of hypervolume and sparsity. The reference points for hypervolume calculation are set to $(0,0)$.}
\label{tab:ftn_exp}
\resizebox{\linewidth}{!}{
\begin{tabular}{@{}lcccccc@{}}
\toprule
 & \multicolumn{2}{c}{Fruit Tree Navigation   (d=5)} & \multicolumn{2}{c}{Fruit Tree Navigation   (d=6)} & \multicolumn{2}{c}{Fruit Tree Navigation   (d=7)} \\ \midrule
 & Hypervolume & Sparsity & Hypervolume & Sparsity & Hypervolume & Sparsity \\\midrule
Envelope & \textbf{6920.58} & N/A & 8427.51 & N/A & 6395.27 & N/A \\
PD-MORL & \textbf{6920.58} & N/A & 9299.15 & N/A & 11419.58 & N/A \\ 
\textbf{PSL-MORL} & \textbf{6920.58} & \textbf{0.01} & \textbf{9302.38} & \textbf{0.01} & \textbf{11786.10}  & N/A
\\ \bottomrule
\end{tabular}
}
\end{table}

\subsection{Results on MORL Benchmarks with Discrete State-Action Spaces}\label{subsec-discrete benchmark}

We evaluate the proposed PSL-MORL on the commonly used benchmark Fruit Tree Navigation~(FTN) with different depths, i.e., $d=5,6,7$.
As shown in Table \ref{tab:ftn_exp}, PSL-MORL achieves the best performance in both metrics in all the environments.
When the depths are $5$ and $6$, Envelope~\cite{yang2019generalized} and PD-MORL~\cite{basaklar2023pdmorl} fail to find the whole Pareto set but find only one policy, resulting in an N/A sparsity value. However, PSL-MORL can find a better approximated Pareto set and get a $0.01$ sparsity. The results demonstrate that PSL-MORL has a better exploration capability than Envelope and PD-MORL. Furthermore, as the depth increases from $5$ to $7$, the hypervolume gap between PSL-MORL and the other two methods also increases, highlighting the greater advantage in more complex environments.

\subsection{Ablation Study}\label{subsec-ablation}

The above experiments have shown the advantage of PSL-MORL, which employs the parameter fusion technique, i.e., $(1-\alpha)\theta_1+\alpha \theta_2$, to merge the parameters from the policy $\theta_2$ generated by the hypernetwork $\phi$ and the initial original policy $\theta_1$. For the choice of $\alpha$, we pick the value that owns the best performance in grid search experiments, as shown in Appendix~A.6. This brings two natural questions: (i) Whether the parameter fusion technique (i.e., setting $\alpha<1$) really improves the performance? (ii) How about adding the two policy parameters (i.e., $\theta_1+ \theta_2$) directly?

To answer the questions, we implement two algorithms on Mo-MuJoCo: (i) PSL-MORL-gen, which uses only the parameters of the policy network generated by the hypernetwork, i.e., without using the parameter fusion technique. (ii) PSL-MORL-add, where the parameters in the policy network are obtained by directly adding the parameters generated by the hypernetwork and the initial policy network. The results are shown in Table~\ref{tab:ablation}. We can observe that when obtaining the parameters entirely from the generation of hypernetwork or directly adding the two portions, the performance of the algorithm will drop drastically, demonstrating the effectiveness of our parameter fusion technique, which truly stabilizes the hypernetwork training.

\vspace{-0.5em}

\section{Conclusion}
In this paper, we introduce the PSL-MORL method, which employs the idea of Pareto Set Learning in MORL and can be viewed as adopting the decomposition idea to deal with all the trade-off preferences. PSL-MORL can allow a decision-maker to make real-time decisions without additional procedures. Experimental results show that PSL-MORL outperforms previous methods, and more specifically, PSL-MORL achieves the best hypervolume and sparsity in most benchmarks. Furthermore, the theoretical results guarantee the superiority of the model capacity and the optimality of the generated policy network for our proposed method. One limitation of this work is that we only consider the linearized reward setting. It would be interesting to research on the non-linear scalarization functions, structure of the hypernetwork and other advanced models, which may bring further performance improvement.

\newpage

\section{Acknowledgements}
The authors want to thank the anonymous reviewers for their helpful comments and suggestions. This work was supported by the National Science and Technology Major Project (2022ZD0116600), the National Science Foundation of China (62276124), and the Fundamental Research Funds for the Central Universities (14380020). Chao Qian is the corresponding author.

\bibliographystyle{plainnat}
\bibliography{aaai25} 

\newpage

\section{A. Experiment Details and Additional Results}
\label{appendix:exp}

In this section, we provide the pseudo codes of PSL-MORL with DDQN and PSL-MORL with TD3 in A.1 and A.2, respectively. The details of training hyperparameter, benchmark and baselines are shown in A.3 to A.5, respectively.

\subsection{A.1 Instantiation: PSL-MORL with DDQN}
\label{appendix:psldqn}

In this subsection, we introduce an instantiation of PSL-MORL by taking DDQN~\citep{van2016deep} as the single-objective RL algorithm. The pseudo code is shown in Algorithm~\ref{example:pslmorl_dqn}.

PSL-MORL with DDQN first initializes the hypernetwork and HER. The parallel child processes are the same as that in Algorithm~\ref{alg_psl_morl}. After that, it operates on a batch of preferences. For each preference, the multi-dimensional interpolator $I(\bm\omega)$ is used to obtain the interpolated preference $\bm w_{p}$, and a modified Q-learning update scheme (with the cosine similarity term $S_c$ proposed in~\citep{basaklar2023pdmorl}) is used to calculate the target value $\bm{y}$. The empirical expected loss is then obtained by calculating the DDQN loss for each preference. Finally, an optimizer (e.g., SGD) is employed to update the parameters of both the hypernetwork and target network.

\begin{algorithm}[t!]
    \caption{Instantiation: PSL-MORL with DDQN}
    \label{example:pslmorl_dqn}
    \textbf{Input:}
        Preference distribution $\Lambda$, environment $\mathcal{E}$, number $E$ of episodes, number $K$ of weights per episode, discount factor $\gamma$, target network update coefficient $\tau$, replay buffer $\mathcal{D}$, batch size $N$, multi-dimensional interpolator $I(\boldsymbol{\omega})$ \\
    \textbf{Output:}
        Hypernetwork parameters $\phi$, primary Q-network parameters $\theta_1$, and target Q-network parameters $\theta_1'$
        \\\vspace{-1em}
    \begin{algorithmic}[1]
    \STATE Initialize the hypernetwork parameters $\phi$, current Q network parameters $\theta_1$ and target Q network parameters $\theta_1^\prime$;
    \vspace{-1em}
    \STATE Initialize replay buffer $\mathcal{D}$;
        \FOR{episode $e = 1$ to $E$}
          \FOR{each step in the episode}
            \STATE // Parallel child processes for $k \in \{1,2,...,K\}$
            \STATE Follow the child process of Algorithm~\ref{alg_psl_morl} (General PSL-MORL);
            \STATE // Main process
            \FORALL{$i\in\{1,2,...,N\}$}
                \STATE Sample transition $(s, a, \bm r, s^\prime)$ from $\mathcal{D}$;
                \STATE $\boldsymbol\omega \sim \textbf{SamplePreference}(\Lambda)$;
                \STATE Use the interpolator $I({\boldsymbol\omega})$ to obtain ${{\boldsymbol\omega}}_p$;
                \STATE Compute the target value $\bm{y}$ using $\bm{r} + \gamma \cdot \mathbf{Q}(s^\prime,\sup_{\substack{a^\prime}}(S_c({{\boldsymbol\omega}}_p,\mathbf{Q}(s^\prime,a^{\prime},{\boldsymbol\omega}))\cdot({\boldsymbol\omega}^{\mathrm{T}}\mathbf{Q}(s^\prime,a^{\prime},{\boldsymbol\omega}))),{\boldsymbol\omega};\theta_1^\prime)$;
                \STATE Generate parameters of the policy network by hypernetwork, i.e., $\theta_{2} =\phi(\boldsymbol{w})$;
                \STATE Obtain the mixed $\theta$ using the \textbf{Parameter Fusion} method on $\theta_1$ and $\theta_{2}$;
                \STATE Compute loss $L_i(\theta)$ using $(\bm{y} - \mathbf{Q}(s,a,{\boldsymbol\omega};\theta))^2$
            \ENDFOR
            \STATE Compute the empirical expected loss $L(\theta)=\sum_{i=1}^N L_i(\theta)/N$;
            \STATE Update $\theta_1$ and $\phi$ to minimize $L(\theta)$;
            \STATE Update target network parameters $\theta_1^\prime$ using $\tau \theta_1 + (1-\tau)\theta_1^\prime$
          \ENDFOR
        \ENDFOR	
    \end{algorithmic}
\end{algorithm}

\subsection{A.2 Instantiation: PSL-MORL with TD3}\label{instantiation:PSL-DDQN for TD3}

In this subsection, we introduce an instantiation of PSL-MORL by taking TD3~\citep{dankwa2019twin} as the single-objective RL algorithm. The pseudo code is shown in Algorithm~\ref{algo:psl_motd3_app}.

The algorithm first initializes the critic network parameters $\varphi_1$, $\varphi_2$, hypernetwork parameters $\phi$, actor network parameters $\theta_1$, target network parameters ${\varphi_1^\prime}$, ${\varphi_2^\prime}$, ${\theta_1^\prime}$, replay buffer $\mathcal D$, and number $n$ of training steps in lines~1--2 of Algorithm~\ref{algo:psl_motd3_app}. For each step in each episode, we first follow Algorithm~\ref{alg_psl_morl} to collect data from the environment in line 6. Then, in the main process, we randomly sample $N$ transitions from the replay buffer and sample $N$ preferences in lines~8--9. In line 10, we utilize the hypernetwork to obtain the generated policy parameters $\theta_2$, and get the mixed target policy parameters $\theta^\prime = (1 - \alpha) \theta_1^\prime + \alpha \theta_2$ using the parameter fusion technique in line 11. After that, we compute the next action $a^\prime$ and obtain $\boldsymbol \omega_p$ by the interpolator in lines~12--13. We then compute the target value $\bm y$ and $g({\boldsymbol\omega}_p,\mathbf{Q}(s,a,\boldsymbol{\omega};\varphi_j))$ in lines~14--15. The critic loss is computed and used to update the critic network parameters in lines~16--17, and the parameters of target critics are updated in line~18. When the critics are trained $p_\text{delay}$ times, we train the actors in lines~21--24, where we get the mixed policy parameters $\theta$ in lines~21--22. The hypernetwork parameter and the actor parameter are updated with policy gradient in line~23, and the target actor parameter is updated in line~24. 

\begin{algorithm}[t!]
    \caption{Instantiation: PSL-MORL with TD3}~\label{algo:psl_motd3_app}
    \textbf{Input:} 
        Preference distribution $\Lambda$, environment $\mathcal{E}$, number $E$ of episodes, number $K$ of weights per episode, discount factor $\gamma$, target network update coefficient $\tau$, replay buffer $\mathcal{D}$, batch size $N$, multi-dimensional interpolator $I(\boldsymbol{\omega})$, policy update delay $p_\text{delay}$, standard deviation $\sigma$ for Gaussian exploration noise added to the policy, standard deviation $\sigma^\prime$ for smoothing noise added to the target policy, limit $c$ for the absolute value of the target policy's smoothing noise \\
    \textbf{Output:}
        Hypernetwork parameters $\phi$, and actor network parameters $\theta_1$
    \begin{algorithmic}[1]
        \STATE Initialize: Critic network parameters $\varphi_1$ and $\varphi_2$, hypernetwork parameters $\phi$, actor network parameters $\theta_1$, and target network parameters $\varphi_1^\prime $, $\varphi_2^\prime$, and $\theta_1^\prime$; 
        \STATE Initialize the replay buffer $\mathcal{D}$ and the number of training steps $n\gets 0$;
        \FOR{episode $e = 1$ to $E$}
            \FOR{each step in the episode}
                \STATE // Parallel child processes for $k \in \{1,2,...,K\}$
                \STATE Follow the child process of Algorithm~\ref{alg_psl_morl} (General PSL-MORL);
                \STATE // Main process
                \STATE Sample transitions $(s, a, \bm r, s^\prime)$ from $\mathcal{D}$;
                \STATE $\boldsymbol\omega \sim \textbf{SamplePreference}(\Lambda)$;
                \STATE Generate parameters of the policy network by hypernetwork, i.e., $\theta_{2} =\phi(\boldsymbol{w})$;
                \STATE Obtain the mixed $\theta^\prime$ using the \textbf{Parameter Fusion} method on $\theta_1^\prime$ and $\theta_{2}$;
                \STATE Select the action $a^\prime$ using $\pi(s^\prime,\boldsymbol{\omega};\theta^\prime) + \epsilon$, where $\epsilon \sim \text{clip}(\mathcal{N}(0,\sigma^\prime),-c,c)$;
                \STATE Use the interpolator $I({\boldsymbol\omega})$ to obtain ${{\boldsymbol\omega}}_p$;
                \STATE Compute the target value $\bm{y}$ using $\bm{r} + \gamma \cdot\arg\min_{\mathbf{Q},j\in\{1,2\}}\boldsymbol{\omega}^{\mathrm{T}}\mathbf{Q}(s^\prime,\sup_{\substack{a^\prime}}(S_c({{\boldsymbol\omega}}_p,\mathbf{Q}(s^\prime,a^{\prime},{\boldsymbol\omega}))\cdot({\boldsymbol\omega}^{\mathrm{T}}\mathbf{Q}(s^\prime,a^{\prime},{\boldsymbol\omega}))),\boldsymbol{\omega};\varphi_j^\prime)$;
                \STATE Compute $g({\boldsymbol\omega}_p,\mathbf{Q}(s,a,\boldsymbol{\omega};\varphi_j))$ using $ \cos^{-1}\frac{{\boldsymbol\omega}_p^{\mathrm{T}} \mathbf{Q}(s,a,\boldsymbol{\omega};\varphi_j)}{\|{\boldsymbol\omega}_p\| \cdot\|\mathbf{Q}(s,a,\boldsymbol{\omega};\varphi_j)\|}$;
                \STATE Compute $L_{\text{critic}}(\varphi_j) = (\bm{y} - \mathbf{Q}(s,a,\boldsymbol{\omega};\varphi_j))^2 + g({\boldsymbol\omega}_p,\mathbf{Q}(s,a,\boldsymbol{\omega};\varphi_j))$;
                \STATE Update $\varphi_1$ and $\varphi_2$ by applying SGD to $L_\text{critic}(\varphi_j)$;
                \STATE Update target critics parameters $\varphi^\prime_{i} \leftarrow \tau \varphi_{i} + (1-\tau)\varphi^\prime_{i}$;
                \STATE $n \gets n + 1$;
                \IF{$n\ \text{mod}\ p_\text{delay} = 0$} 
                \STATE Generate parameters of the policy network by hypernetwork, i.e., $\theta_{2} =\phi(\boldsymbol{w})$;
                \STATE Obtain the mixed $\theta$ using the \textbf{Parameter Fusion} method on $\theta_1$ and $\theta_{2}$;
                \STATE Update $\theta_1$ and $\phi$ with the gradient $\nabla_\theta L_{\text{actor}}(\theta) = \nabla_a~\boldsymbol{\omega}^{\mathrm{T}}\mathbf{Q}(s,a,\boldsymbol{\omega};\varphi_1)|_{a=\pi(s,\boldsymbol{\omega};\theta)}\nabla_{\theta} \pi(s,\boldsymbol{\omega};\theta) + \alpha \cdot \nabla_a~g({\boldsymbol\omega}_p,\mathbf{Q}(s,a,\boldsymbol{\omega};\varphi_1))|_{a=\pi(s,\boldsymbol{\omega};\theta)}\nabla_{\theta} \pi(s,\boldsymbol{\omega};\theta)$;
                
                \STATE Update target actor parameters $\theta_1^\prime \leftarrow \tau \theta_1 + (1-\tau)\theta_1^\prime$
                \ENDIF
            \ENDFOR
        \ENDFOR
    \end{algorithmic}
\end{algorithm}

\subsection{A.3 Training Details}
\label{appendix:training_detail}
The hyperparameter settings of PSL-MORL with DDQN and PSL-MORL with TD3 are shown in Tables~\ref{tab:mo_ddqn_her_hyperparameters} and~\ref{tab:mo_td3_her_hyperparameters}, respectively. The shared hyperparameters, except the parameter fusion coefficient, are all the same as that of previous baseline algorithms. The experiment of grid search on parameter fusion coefficient is provided in Appendix~A.6.

\subsection{A.4 Benchmark Details}
\label{appendix:benchmarkdetails}
We show the benchmark details on MO-MuJoCo and Fruit Tree Navigation (FTN) in the experiments. The objectives, and the dimensions of state and actions are shown in Table~\ref{tab:mujoco-setiings}. We include five different environments of MO-MuJoCo in our experiment, including MO-HalfCheetah-v2, MO-Hopper-v2, MO-Ant-v2, MO-Swimmer-v2, and MO-Walker-v2.

\begin{itemize}

\item\textbf{MO-HalfCheetah-v2:} The agent is a bidimensional robotic entity resembling a cheetah, tasked with optimizing two goals: maximization of forward momentum and conservation of energy. The environments for state and action are defined as $S \subseteq \mathbb{R}^{17}, A \subseteq \mathbb{R}^{6}$. The aim is to adjust the torque on the limb joints according to a preference vector $\bm{\omega}$ to achieve efficient forward motion.

\item\textbf{MO-Hopper-v2:} With the state and action spaces described as $S\subseteq \mathbb{R}^{11}, A\subseteq \mathbb{R}^{3}$, the agent takes the form of a bidimensional mono-legged robot. It focuses on two main objectives: the acceleration of forward movement and the maximization of jump height. Adjusting the torque on its hinge according to a preference vector $\bm{\omega}$ is key to its forward hopping motion.

\item\textbf{MO-Ant-v2:} The agent is represented as a three-dimensional robotic ant, aiming to balance two objectives: speed along the $x$-axis and speed along the $y$-axis. Its main goal is to adjust the torque on the leg and torso connectors in line with a preference vector $\bm{\omega}$. The defined state and action spaces are $S\subseteq \mathbb{R}^{27}, A\subseteq \mathbb{R}^{8}$. 

\item\textbf{MO-Swimmer-v2:} The task trains a bidimensional robotic entity to optimize the forward velocity and the energy efficiency as it swims in a two-dimensional pool. This benchmark defines state and action spaces as $S\subseteq \mathbb{R}^{8}, S\subseteq \mathbb{R}^{2}$. The primary objective is to fine-tune the torque on its rotors according to a preference vector $\bm{\omega}$.

\item\textbf{MO-Walker-v2:} The agent, a bidimensional bipedal robot, seeks to achieve two goals: enhancement of forward speed and energy conservation. The challenge lies in adjusting the torque on its hinges to align with a preference vector $\bm{\omega}$ while moving forward. The environment's state and action spaces are $S\subseteq \mathbb{R}^{17}, A\subseteq \mathbb{R}^{6}$.

\item
\textbf{Fruit Tree Navigation (FTN):} An innovative MORL benchmark introduced by ~\citep{yang2019generalized}. It involves a binary tree of depth $d$, where each leaf node is assigned a reward vector $\bm r\in \mathbb{R}^6$ representing the nutritional values of fruits, categorized into six types: {Protein, Carbs, Fats, Vitamins, Minerals and Water}. The goal of the agent is to navigate the tree to harvest fruit that optimizes these nutritional values based on specific preferences.
\end{itemize}

\begin{table*}[t!]
\caption{Hyperparameter settings of PSL-MORL with DDQN.}
\label{tab:mo_ddqn_her_hyperparameters}
\resizebox{\textwidth}{!}{
\small
\centering
\begin{tabular}{@{}lccc@{}}
\toprule

\textbf{}  & \multicolumn{1}{l}{\textbf{Fruit Tree Navigation ($d=5$)}} & \multicolumn{1}{l}{\textbf{Fruit Tree Navigation ($d=6$)}} & \multicolumn{1}{l}{\textbf{Fruit Tree Navigation ($d=7$)}}\\ \midrule
\textbf{Total number of steps}  & $1\times10^5$ & $1\times10^5$ & $1\times10^5$\\
\textbf{Minibatch size}  & 32 & 32 & 32\\
\textbf{Discount factor}  & 0.99 & 0.99 & 0.99\\
\textbf{Soft update coefficient} & 0.005 & 0.005 & 0.005\\
\textbf{Buffer size} & $1\times10^4$ & $1\times10^4$ & $1\times10^4$ \\
\textbf{Number of child processes }  & 10 & 10 & 10 \\
\textbf{Number of preferences sampled for HER } & 3  & 3  & 3\\
\textbf{Parameter fusion coefficient}  &  0.05 & 0.05 & 0.10   \\
\textbf{Learning rate}  & $3\times10^{-4}$ & $3\times10^{-4}$ & $3\times10^{-4}$\\
\textbf{Number of hidden layers}  & 3 & 3 & 3\\
\textbf{Number of hidden neurons}  & 512 & 512 & 512\\ 
     \bottomrule
\end{tabular}
}
\end{table*}

\begin{table*}[t!]
\caption{Hyperparameter settings of PSL-MORL with TD3.}
\label{tab:mo_td3_her_hyperparameters}
\resizebox{\textwidth}{!}{
\begin{tabular}{@{}lccccc@{}}
\toprule
\textbf{} & \multicolumn{1}{l}{\textbf{MO-Walker2d-v2}} & \multicolumn{1}{l}{\textbf{MO-HalfCheetah-v2}} & \multicolumn{1}{l}{\textbf{MO-Ant-v2}} & \multicolumn{1}{l}{\textbf{MO-Swimmer-v2}} & \multicolumn{1}{l}{\textbf{MO-Hopper-v2}} \\ \midrule
\textbf{Total number of steps} & $1\times10^6$ & $1\times10^6$ & $1\times10^6$ & $1\times10^6$ & $1\times10^6$ \\
\textbf{Minibatch size} & 256 & 256 & 256 & 256 & 256 \\
\textbf{Discount factor} & 0.995 & 0.995 & 0.995 & 0.995 & 0.995 \\
\textbf{Soft update coefficient} & 0.005 & 0.005 & 0.005 & 0.005 & 0.005 \\
\textbf{Buffer size} & $2\times10^6$ & $2\times10^6$ & $2\times10^6$ & $2\times10^6$ & $2\times10^6$ \\
\textbf{Number of child processes} & 10 & 10 & 10 & 10 & 10 \\
\textbf{\begin{tabular}[c]{@{}l@{}}Number of preferences \\ sampled for HER\end{tabular}} & 3 & 3 & 3 & 3 & 3 \\ 
\textbf{Learning rate - Critic} & $3\times10^{-4}$ & $3\times10^{-4}$ & $3\times10^{-4}$ & $3\times10^{-4}$ & $3\times10^{-4}$ \\
\textbf{Number of hidden layers -   Critic} & 1 & 1 & 1 & 1 & 1 \\
\textbf{Number of hidden neurons -   Critic} & 400 & 400 & 400 & 400 & 400 \\ 
\textbf{Learning rate - Actor} & $3\times10^{-4}$ & $3\times10^{-4}$ & $3\times10^{-4}$ & $3\times10^{-4}$ & $3\times10^{-4}$ \\
\textbf{Number of hidden layers -   Actor} & 1 & 1 & 1 & 1 & 1 \\
\textbf{Number of hidden neurons -   Actor} & 400 & 400 & 400 & 400 & 400 \\
\textbf{Policy update delay} & 10 & 10 & 10 & 10 & 20 \\
\textbf{Parameter fusion coefficient} & 0.01 & 0.05 & 0.05 & 0.01 & 0.03 \\
\textbf{Exploration noise std.} & 0.1 & 0.1 & 0.1 & 0.1 & 0.1 \\
\textbf{Target policy's smoothing noise std.} & 0.2 & 0.2 & 0.2 & 0.2 & 0.2 \\
\textbf{Noise clipping limit} & 0.5 & 0.5 & 0.5 & 0.5 & 0.5 \\
\textbf{Loss coefficient} & 10 & 10 & 10 & 10 & 10 \\ 
 \bottomrule
\end{tabular}
}
\end{table*}

\begin{table*}[t!]
    \centering
    \caption{Detailed objectives, and dimensions of state and action of MO-MuJoCo benchmarks. All these five benchmarks are continuous robotic control tasks. The number of objectives for these benchmarks all equals two. }
    \label{tab:mujoco-setiings}
    \begin{tabular}{c|ccc}
        \toprule
        Environments & Dim of $S$ & Dim of $A$ & Objectives \\ \midrule
        MO-Walker-v2 & 17 & 6 & forward speed, energy efficiency \\
        MO-HalfCheetah-v2 & 17 & 6 & forward speed, energy efficiency \\
        MO-Ant-v2 & 27 & 8 & $x$-axis speed, $y$-axis speed \\
        MO-Swimmer-v2 & 8 & 2 & forward speed, energy efficiency \\ 
        MO-Hopper-v2 & 11 & 3 & forward speed, jumping height \\
        \bottomrule
    \end{tabular}
\end{table*}

\subsection{A.5 Baselines Details}
\label{appendix:baselines}
The detailed descriptions of MORL baselines compared in the experiments are as follows:
\begin{itemize}
\item\textbf{PG-MORL}~\citep{xu2020prediction}: PG-MORL uses a prediction-guided evolutionary learning method to find high-quality solutions and conducts Pareto analysis to get the ultimate Pareto front.
\item\textbf{PD-MORL}~\citep{basaklar2023pdmorl}: PD-MORL utilizes the preference as guidance to update the network parameters and covers the entire preference space.
\item\textbf{Radial Algorithm (RA)}~\citep{parisi2014policy}: The method assigns a set of weights to scalarize the objectives and runs single objective reinforcement learning algorithms to train a policy for each weight separately.
\item\textbf{Pareto Front Adaption (PFA)}~\citep{parisi2014policy}: The method fine-tunes the optimized weight to cover the whole Pareto front. It makes transfers from some weights to their neighborhoods, but the transferred policy may not be optimal.
\item\textbf{MOEA/D}~\citep{moead}: The method decomposes a multi-objective optimization problem into a number of scalar optimization subproblems, and then optimizes them collaboratively. 
\item\textbf{RANDOM}: A variant of PG-MORL that uniformly samples weights in each generation.
\item\textbf{META}~\citep{chen2019meta}: The method trains a meta policy and adapts it to different weights.
\end{itemize}

\subsection{A.6 Influence of Parameter Fusion Coefficient}\label{sec-appendix-grid-search}
In this section, we present the experimental results about the performance of using different coefficients in parameter fusion in Table~\ref{tab:ablation_exp}. The bold numbers are the best within each column. We can observe that for MO-Walker2d-v2, MO-HalfCheetah-v2, MO-Ant-v2 and MO-Swimmer-v2, the best parameter fusion coefficient is $\alpha=0.01$; while for MO-Hopper-v2, the best parameter fusion coefficient is $\alpha=0.03$. Utilizing the parameter fusion can leverage both the hypernetwork and primary network to make the trained policy more robust.

\begin{table*}[t!]
\caption{Grid search on parameter fusion experiment.}
\resizebox{\textwidth}{!}{
\begin{tabular}{@{}lcccccccccc@{}}
\toprule
 & \multicolumn{2}{c}{\textbf{MO-Walker2d-v2}} & \multicolumn{2}{c}{\textbf{MO-HalfCheetah-v2}} & \multicolumn{2}{c}{\textbf{MO-Ant-v2}} & \multicolumn{2}{c}{\textbf{MO-Swimmer-v2}} & \multicolumn{2}{c}{\textbf{MO-Hopper-v2}}  \\ \midrule
  $\bm{\alpha}$ & \multicolumn{1}{c}{\textbf{HV $(\times10^6)$}} & \multicolumn{1}{c}{\textbf{SP $(\times10^4)$}} & \multicolumn{1}{c}{\textbf{HV} $(\times10^6)$} & \multicolumn{1}{c}{\textbf{SP $(\times10^3)$}} & \multicolumn{1}{c}{\textbf{HV} $(\times10^6)$} & \multicolumn{1}{c}{\textbf{SP $(\times10^4)$}} & \multicolumn{1}{c}{\textbf{HV} $(\times10^4)$} & \multicolumn{1}{c}{\textbf{SP $(\times10)$}} & \multicolumn{1}{c}{\textbf{HV} $(\times10^7)$} & \multicolumn{1}{c}{\textbf{SP $(\times10^4)$}}  \\\midrule

$0.01$ & $\mathbf{5.36}$ & $\mathbf{0.01}$ & $\mathbf{5.92}$ & $\mathbf{0.16}$ & $\mathbf{8.63}$ & $\mathbf{0.68}$ & $\mathbf{3.22}$ & $0.42$ & $1.77$ & $0.99$ \\

$0.03$ & $5.28$ & $0.01$ & $5.91$ & $0.29$ & $7.23$ & $0.94$ & $3.21$ & $1.02$ & $\mathbf{1.95}$ & $0.68$ \\
  
$0.05$ & $4.82$ & $0.02$ & $5.91$ & $0.20$ & $7.20$ & $1.24$ & $3.17$ & $0.77$ & $1.84$ & $0.87$  \\

$0.10$
& $4.93$   & $0.03$ & $5.90$  &  $0.30$ & $7.45$  &  $0.96$ &  $3.16$   &   $0.99$  &  $1.77$ & $0.97$    \\

$0.15$
& $4.83$ & $0.14$ & $5.87$ & $0.60$ & $7.47$ & $0.88$ & $3.11$ & $1.09$ & $1.85$ & $1.45$ \\

$0.20$
& $4.19$ & $0.09$ & $5.89$ & $0.45$ & $7.04$ & $0.94$ & $3.12$ & $0.74$ & $1.67$ & $0.58$ \\

$0.25$
& $4.48$ & $0.06$ & $5.88$ & $0.32$ & $6.82$ & $1.08$ & $3.06$ & $0.69$ & $1.56$ & $2.05$ \\

$0.30$
& $4.56$ & $0.04$ & $5.78$ & $0.74$ & $5.64$ & $1.23$ & $3.09$ & $\mathbf{0.36}$ & $1.47$ & $0.62$ \\

$0.35$
& $3.96$ & $0.16$ & $5.80$ & $0.33$ & $5.88$ & $1.28$ & $3.19$ & $1.01$ & $1.53$ & $0.74$ \\

$0.40$
& $3.15$ & $0.26$ & $5.72$ & $0.53$ & $4.99$ & $1.18$ & $3.06$ & $0.54$ & $1.48$ & $1.97$ \\

$0.45$
& $2.91$ & $0.22$ & $5.77$ & $0.36$ & $3.92$ & $1.08$ & $3.08$ & $1.09$ & $1.45$ & $\mathbf{0.37}$ \\

$0.50$
& $2.84$ & $0.03$ & $5.59$ & $0.54$ & $3.17$ & $0.72$ & $3.07$ & $0.92$ & $1.44$ & $1.29$ \\

\bottomrule
\end{tabular}
}
\label{tab:ablation_exp}
\end{table*}

\section{B. Proof of Theorem~\ref{thm:rademacher}}
\label{appendix:proofthm2}
In this section, we provide the complete proof for Theorem~\ref{thm:rademacher}. The proof relies on the existing results on Rademacher complexity of MLPs. We first provide some useful facts and definitions. 

\subsection{B.1 Preliminary}    
Given a vector $\mathbf{w} \in \mathbb{R}^h$, let $\lVert \mathbf{w} \rVert$ denote the Euclidean norm. For $p \geq 1$, $\lVert \mathbf{w} \rVert_p = \left( \sum_{i=1}^h |w_i|^p \right)^{1/p}$ denotes the $\ell_p$ norm. For a matrix $W$, we denote $\lVert W \rVert_p$, where $p \in [1,\infty]$, as the Schatten $p$-norm (i.e., the $p$-norm of the spectrum of $W$). For instance, $p=\infty$ corresponds to the spectral norm, $p=2$ to the Frobenius norm, and $p=1$ to the trace norm. In the case of the spectral norm, we omit the $\infty$ subscript and simply use $\lVert W \rVert$. Additionally, we use $\lVert W \rVert_F$ to represent the Frobenius norm. 

Considering the domain $\mathcal{X} = \{\mathbf{x} : \lVert \mathbf{x} \rVert \leq B\}$ in Euclidean space, we examine standard neural networks, which can be scalar or vector-valued, and are structured as follows:
\[
\mathbf{x} \mapsto W_d \sigma_{d-1}(W_{d-1} \sigma_{d-2}(\ldots \sigma_1(W_1 \mathbf{x}))),
\]
where each $W_j$ represents a parameter matrix, and each $\sigma_j$ is a fixed Lipschitz continuous function mapping between Euclidean spaces, with the property that $\sigma_j(\mathbf{0}) = \mathbf{0}$. The variable $d$ signifies the depth of the network, while its width $h$ is determined by the maximum of the row or column dimensions of the matrices $W_1, \ldots, W_d$. For simplicity, we assume that each $\sigma_j$ has a Lipschitz constant no greater than $1$; if not, the Lipschitz constant can be incorporated into the norm constraint of the adjacent parameter matrix. We classify $\sigma$ as element-wise if it applies the same univariate function to each coordinate of its input. $\sigma$ is termed positive-homogeneous if it is element-wise and fulfills the condition $\sigma(\alpha z) = \alpha \sigma(z)$ for all $\alpha \geq 0$ and $z \in \mathbb{R}$.

\begin{definition}(Rademacher Complexity, Definition 13.1 in~\citep{golowich18a})
    Given a real-valued function class $\Hcal$ and some set of data 
points 
$\bx_1,\ldots,\bx_n\in \Xcal$, we define the (empirical) Rademacher 
complexity 
$\hat{\Rcal}_n(\Hcal)$ as 
\begin{equation}\label{eq:raddef}
\hat{\Rcal}_n(\Hcal) = 
\mathbb{E}_{\boldsymbol{\epsilon}}\left[\sup_{h\in\Hcal}\frac{1}{n}\sum_{i=1}^{n}
\varepsilon_i h(\bx_i)\right],
\end{equation}
where $\boldsymbol{\varepsilon}=(\varepsilon_1,\ldots,\varepsilon_n)$ is a 
vector 
uniformly distributed in $\{-1,+1\}^n$.
\end{definition}

\subsection{B.2 Rademacher Complexity of Neural Networks}
In this section, we present an upper bound and a lower bound on the Rademacher complexity, given by ~\citep{golowich18a}, for the class of neural networks with parameter matrices of bounded Schatten norms. They will be used in our proof.

Assuming that $\norm{\mathbf{w}}\leq M$ (where $\norm{\cdot}$	signifies 	Euclidean norm), and the sample distribution satisfies that $\norm{\bx}\leq B$ almost surely (which holds with probability $1$). That is, $B$ is the upper bound on the norm of all the sample points.

\begin{lemma}~\citep{golowich18a}
\label{thm:sqrtl}
Let $\mathcal{H}_d$ be the class of real-valued networks of depth $d$ over 
the domain $\mathcal{X}$, where  each parameter matrix $W_j$ has Frobenius norm at 
most $M_F(j)$, and the activation function $\sigma$ is a $1$-Lipschitz, positive-homogeneous activation function which is applied element-wise (such as the ReLU). Then,
\begin{align*}
\hat{\Rcal}_n(\Hcal_d)
&\leq \frac 1n \prod_{j=1}^d M_F(j) \cdot  
\left(\sqrt{2\log(2)d}+1\right)\sqrt{\sum_{i=1}^n \|\mathbf{x}_i\|^2}\\ 
&\leq
\frac{B\left(\sqrt{2\log(2)d}+1\right)\prod_{j=1}^d 
    M_F(j)}{\sqrt n}.
\end{align*}
\end{lemma}

\begin{lemma}~\citep{golowich18a}
\label{thm:lowerbound}
Let $\mathcal{H}_d$ be the class of depth-$d$ neural networks over the domain $\mathcal{X}$, where each parameter matrix $W_j$ satisfies $\norm{W_j}_p\leq M_p(j)$ for some Schatten $p$-norm $\norm{\cdot}_p$, i.e., the parameter matrices $W_1,\ldots,W_d$ in the $d$ layers have Schatten norms $\norm{\cdot}_p$ upper-bounded by $M_p(1),\ldots,M_p(d)$, respectively. Then there exists a
$\frac{1}{\gamma}$-Lipschitz loss $\ell$ and data points 
$\bx_1,\ldots,\bx_n\in\mathcal{X}$, such that
\[
\hat{\mathcal{R}}_n(\ell\circ\mathcal{H}_d)~\geq~    \Omega\left(
\frac{B\prod_{j=1}^{d}M_p(j)\cdot 
    h^{\max\left\{0,\frac{1}{2}-\frac{1}{p}\right\}}}{\gamma\sqrt{n}}\right).
\]
\end{lemma}

\subsection{B.3 Proof of Theorem~\ref{thm:rademacher}.}
We are now ready for the complete proof.\vspace{0.5em}

\begin{proof}
    Suppose the class of neural networks for PD-MORL is $\Hcal_{d_1}$ with depth-$d_1$. The additional part in our proposed PSL-MORL method is the hypernetwork, which we assume is a class of neural networks $\Hcal_{d_2}$ with depth $d_2$. In order to avoid confusion, the activation function and parameter matrix for the hypernetwork will be expressed as the form like $W_i'$ and $\sigma_i'$, whereas the activation function and parameter matrix for the neural networks in PD-MORL will be expressed as the form like $W_i$ and $\sigma_i$.
    
    With the above notation for neural networks, we can derive the form of PD-MORL as follows:
    \[
	\bx\mapsto W_{d_1}\sigma_{d_1-1}(W_{d_1-1}\sigma_{d_1-2}(\ldots \sigma_1(W_1\bx))).
	\]
 Similarly, we can obtain the form of PSL-MORL as: 
\begin{align*}
\bx\mapsto (W_{d_2}'\sigma_{d_2-1}'(W_{d_2-1}'\sigma_{d_2-2}'(\ldots \sigma_1'( W_1'))))\circ(W_{d_1-1}\sigma_{d_1-2}(\ldots \sigma_1(W_1\bx))).        
\end{align*}
By Lemma~\ref{thm:lowerbound} with $p=2$, we can obtain the lower bound $\hat{\mathcal{R}}_2$ for PSL-MORL as follows:
    \[
    \hat{\mathcal{R}}_n(\ell_1\circ\mathcal{H}_{d_1-1})~\geq~    \Omega\left(
    \frac{B\prod_{j=1}^{d_1-1}M_2(j)}{\gamma\sqrt{n}}\right)=\hat{\mathcal{R}}_{2,1},
    \]

    \[
    \hat{\mathcal{R}}_n(\ell_2\circ\mathcal{H}_{d_2})~\geq~    \Omega\left(
    \frac{B\prod_{j=1}^{d_2}M'_2(j)}{\gamma\sqrt{n}}\right)=\hat{\mathcal{R}}_{2,2},
    \]

    \[
    \hat{\mathcal{R}}_2 = \hat{\mathcal{R}}_{2,1} \cdot \hat{\mathcal{R}}_{2,2},    
    \]
where $M_2(j)$ and $M'_2(j)$ denote the Schatten norm upper bound as presented in Lemma~\ref{thm:lowerbound}. Because the Schatten norm $p=2$ corresponds to the Frobenius norm, we have
    \[
    \exists \text{ constant } c_1 > 0, \text{s.t. } \hat{\mathcal{R}}_{2,1} \ge c_1 \cdot \frac{B\prod_{j=1}^{d_1-1}M_F(j)}{\gamma\sqrt{n}},    
    \]

    \[
    \exists \text{ constant } c_2 > 0, \text{s.t. } \hat{\mathcal{R}}_{2,2} \ge c_2 \cdot \frac{B\prod_{j=1}^{d_2}M'_F(j)}{\gamma\sqrt{n}},    
    \]

    \[
    \hat{\mathcal{R}}_2 \ge c_1 \cdot c_2 \cdot \frac{B^2\prod_{j=1}^{d_1-1}M_F(j)\prod_{j=1}^{d_2}M'_F(j)}{\gamma^2\cdot n},
    \]
where $M_F(j)$ and $M'_F(j)$ denote the Frobenius norm upper bound as presented in Lemma~\ref{thm:sqrtl}. Furthermore, by Lemma~\ref{thm:sqrtl}, we can get the upper bound for PD-MORL:
    \[
    \hat{\mathcal{R}}_n(\mathcal{H}_{d_1}) \leq 
    \frac{B\left(\sqrt{2\log(2)d_1}+1\right)\prod_{j=1}^{d_1}
        M_F(j)}{\sqrt n}=\hat{\mathcal{R}}_1.
    \]
Taking the ratio of $\hat{\mathcal{R}}_2$ versus $\hat{\mathcal{R}}_1$ leads to

    \[
    \frac{\hat{\mathcal{R}}_2}{\hat{\mathcal{R}}_1} \ge c_1 \cdot c_2 \cdot \frac{B\cdot \prod_{j=1}^{d_2}M'_F(j)}{\gamma^2 \cdot \sqrt{n} \left(\sqrt{2\log(2)d_1}+1\right)M_F(d_1)}.
    \]
Let $M'_F(j)=\max\left\{\|W'_j\|_2,\left(\frac{\gamma^2 \cdot \sqrt{n} \left(\sqrt{2\log(2)d_1}+1\right)M_F(d_1)}{c_1\cdot c_2\cdot B}\right)^{\frac{1}{d_2}}\right\}+1$, which is an obvious upper bound on $\|W_j\|_2$. As $M'_F(j)> \left(\frac{\gamma^2 \cdot \sqrt{n} \left(\sqrt{2\log(2)d_1}+1\right)M_F(d_1)}{c_1\cdot c_2\cdot B}\right)^{\frac{1}{d_2}}$, we can come to the conclusion that ${\hat{\mathcal{R}}_2}/{\hat{\mathcal{R}}_1} > 1$, and complete the proof.    
\end{proof}

\end{document}